\documentclass[conference,10pt]{IEEEtran}
\usepackage{amsmath,graphicx}
\usepackage{url}

\usepackage{amsfonts}
\usepackage{amssymb}
\usepackage{amsbsy}
\usepackage{times}
\usepackage{default}
\usepackage{color}
\usepackage[noadjust]{cite}
\usepackage{enumerate}
\usepackage{tikz}
\usepackage{subcaption}
\usepackage{amsthm}
\usepackage{dsfont}
\usetikzlibrary{arrows}

\usepackage{multirow}

\newcommand{\la}{\lambda}
\newcommand{\R}{\mathbb{R}}

\newtheorem{remark}{Remark}

\newcommand{\rev}[1]{#1}

\usepackage{footnote}
\makesavenoteenv{tabular}
\makesavenoteenv{table}
\usepackage{pgf,tikz}

\usepackage{pgfplots}
\pgfplotsset{compat=1.3}

\usepackage{caption}
\usepackage{subcaption}

\title{\hspace{-0.7mm}Deep Structured Features for Semantic Segmentation}
\author{
\IEEEauthorblockN{Michael Tschannen, Lukas Cavigelli, Fabian Mentzer, Thomas Wiatowski, and Luca Benini}
\IEEEauthorblockA{Dept. IT \& EE, ETH Zurich, Zurich, Switzerland \\
\{michaelt, withomas\}@nari.ee.ethz.ch, mentzerf@student.ethz.ch, \{cavigelli, lbenini\}@iis.ee.ethz.ch}
}

\begin{document}

\maketitle
\begin{abstract}
\rev{We propose a highly structured neural network architecture for semantic segmentation with an extremely small model size, suitable for low-power embedded and mobile platforms. Specifically, our architecture combines i) a Haar wavelet-based tree-like convolutional neural network (CNN), ii) a random layer realizing a radial basis function kernel approximation, and iii) a linear classifier.} While stages i) and ii) are completely pre-specified, only the linear classifier is learned from data. We apply the proposed architecture to outdoor scene and aerial image semantic segmentation and show that the accuracy of our architecture is competitive with conventional pixel classification CNNs. Furthermore, we demonstrate that the proposed architecture is data efficient in the sense of matching the accuracy of pixel classification CNNs when trained on a much smaller data set.
\end{abstract}

\section{Introduction}
\label{sec:intro}

Semantic segmentation of images is an important step in many computer vision applications and refers to the task of identifying the semantic category of every pixel of an image. For example, in images depicting street scenes, the list of possible categories may include ``car'', ``person'', or ``tree''.

In recent years, convolutional neural networks (CNNs) have emerged as a popular method for semantic segmentation. To infer a dense labeling, most of the corresponding CNN architectures either rely on pixel classification \cite{farabet2013learning, paisitkriangkrai2015effective, pinheiro2014recurrent, volpi2016dense} or on so-called deconvolution layers \cite{long2015fully, eigen2015predicting, liang2015semantic, badrinarayanan2015segnet, cavigelli2015accelerating, marmanisa2016semantic, volpi2016dense}, which combine up-sampling, interpolation with a (possibly learned) kernel, and a non-linearity.  
Pixel classification approaches map each pixel to a feature vector by feeding the patch surrounding the pixel through a CNN and then applying a (respectively, the same) classifier to each feature vector. The so-obtained labeling is often refined using, e.g., superpixel and/or conditional random field (CRF)-based regularization. Deconvolution layer-based CNNs construct the labels by connecting deconvolution layers (with the output size being equal to the size of the original image) to one (in the case of encoder-decoder architectures \cite{badrinarayanan2015segnet, volpi2016dense}) or multiple intermediate layers of a feed-forward CNN, the output of these deconvolution layers being finally combined. In both cases, the architectures are typically trained end-to-end. Deconvolution layer-based approaches are often more accurate and faster than pixel classification networks.

Both types of 
architectures have a rather complex structure resulting in large models, and may therefore not be suited for applications subject to strong resource constraints as present in embedded vision systems. 
Furthermore, the necessity of pre-training \cite{long2015fully, eigen2015predicting, liang2015semantic, marmanisa2016semantic}, training on large labeled data sets, and parameter optimization requiring gradient back-propagation through the entire network, 
may hinder on-device learning \cite{shanbhag2016energy} and applications 
where only a small set of labeled images is available. 
\paragraph*{Contributions} We propose a novel, highly structured CNN architecture for semantic segmentation. Specifically, this architecture combines a tree-like CNN-based  feature extractor  \cite{wiatowski2016}, a random layer realizing a radial basis function (RBF) kernel approximation \cite{rahimi2007random}, and a linear classifier \cite{bottou2010large}. 
The feature extractor, typically the computational bottleneck in CNNs, allows for a very fast implementation---in particular when employed with separable wavelet filters as in our experiments. 
All trainable parameters of our architecture are in the last layer (i.e., the linear classifier), allowing for very simple stochastic gradient descent (SGD) weight  updates, well suited for on-device learning. \rev{This is in} contrast to conventional CNN architectures for which SGD weight updates typically require computationally more demanding gradient back-propagation.
Using Haar wavelets as convolutional filters, we evaluate the architecture for semantic segmentation of two different image types, namely outdoor scene images (from the Stanford Background data set \cite{gould2009decomposing}) and aerial images (form the Vaihingen data set of the ISPRS 2D semantic labeling contest \cite{isprsdataset}), using identical values for almost all hyper-parameters of the architecture in both cases. For both image types, the performance of our architecture is \rev{comparable to that of end-to-end trained} conventional pixel classification CNNs. However, our architecture has a model size \rev{of less than 350kBytes, which is 2-3 orders of magnitude smaller than the size of most conventional CNN models}, and is therefore an ideal choice for inference on platforms with strong memory constraints. Further experiments show that our architecture is data efficient and matches the accuracy of the CNN in \cite{farabet2013learning} using a much smaller training set.

\begin{figure}\vspace{1.15cm}
\setlength{\unitlength}{0.8cm}
\begin{picture}(10,5)

\put(.55,3.9){\footnotesize$
f$}

\put(.75,5.65){\footnotesize$
u_{q_1}\ast \chi_1$}

\put(.75,2.4){\footnotesize$
f_{i,j}$}

\put(.75,2.4){\footnotesize$
f_{i,j}$}

\put(8.35,.95){\footnotesize$
F_{i,j}$}

\put(9.1,.4){\footnotesize$
\phi(F_{i,j})$}

\put(10.3,2.35){\footnotesize$
y_{i,j}$}

\put(5.45,4.95){\footnotesize$U_{2}$}
\put(5.45,2.05){\footnotesize$U_{2}$}
\put(5.45,3.5){\footnotesize$U_{2}$}

\put(0,0.75){\includegraphics[width = .47\textwidth]{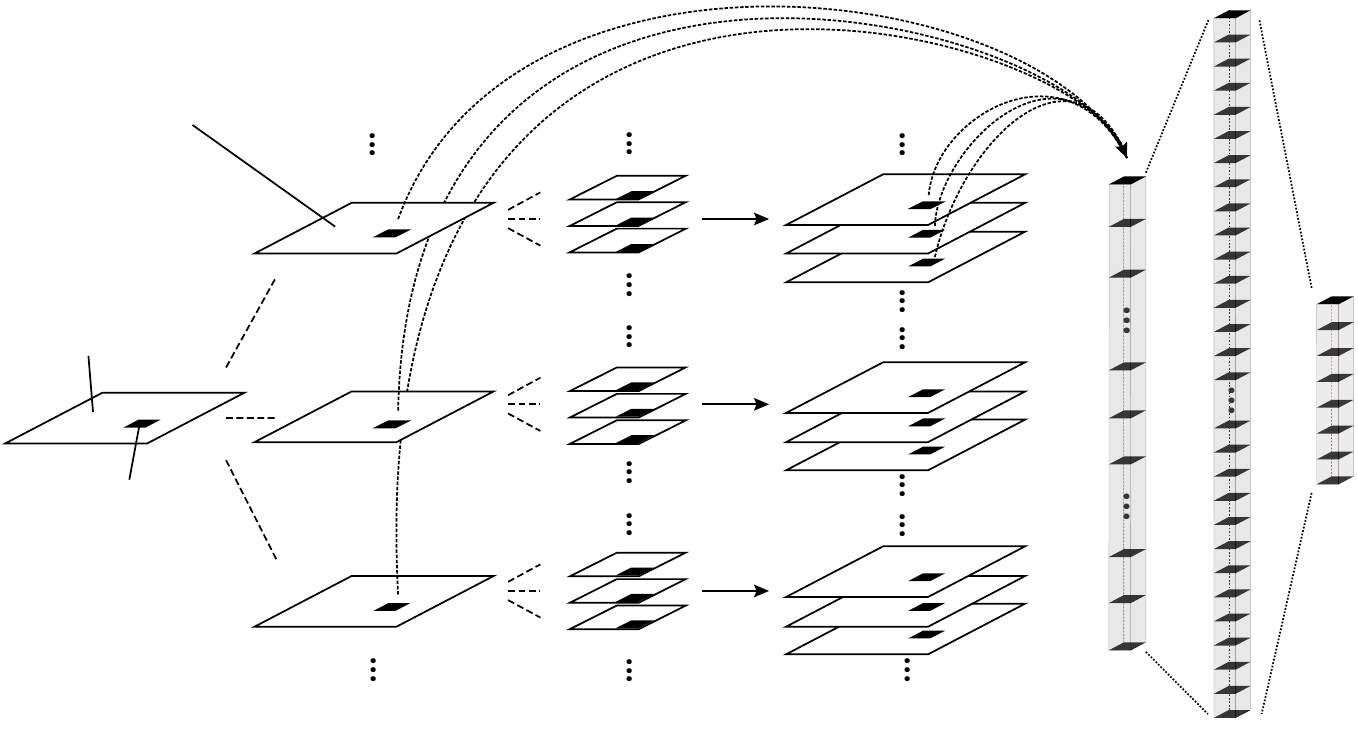} }

\put(0.,1){$\underbrace{\hspace{4.3cm}}$}
\put(6.,1){$\underbrace{\hspace{1.6cm}}$}

\put(1.3,.3){\footnotesize tree-like deep CNN}
\put(5.5,.3){\footnotesize up-sampled feat. maps}

\end{picture}
\vspace{-0.1cm}
\caption{Network architecture for semantic segmentation of total depth $4$. The first $D=2$ layers correspond to the tree-like CNN defined in Sec. \ref{sec:tree_cnn}, which employs pooling by sub-sampling from the first to the second layer with pooling factor 2. The feature vector $F_{i,j}\in \mathbb{R}^m$ corresponding to the pixel $f_{i,j}\in \mathbb{R}$ is transformed to $\phi(F_{i,j})$ according to \eqref{eq:rbf_sam}, and class scores $y_{i,j}\in \mathbb{R}^K$ are computed as in \eqref{eq:svm}.
}\vspace{-0.3cm}
\label{fig_net}
\end{figure}
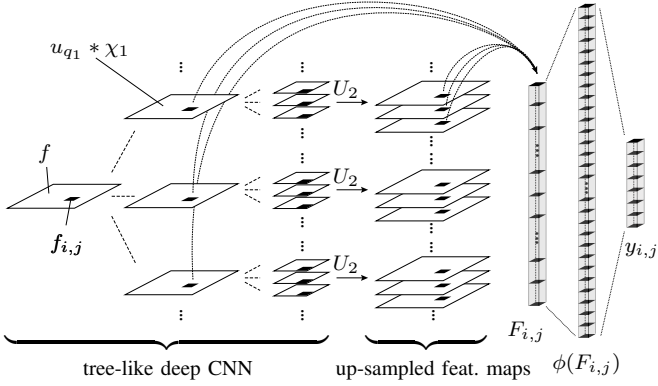

\section{Network architecture}

We set the stage by introducing our CNN architecture for semantic segmentation. The CNN we consider has a total depth of $D+2$ where the first $D$ layers correspond to a tree-like CNN-based feature extractor with pre-specified (i.e., hand-crafted) frame filters \cite{wiatowski2016}, followed by a non-linear classifier composed of a single fully connected RBF kernel approximation layer with pre-specified random filters \cite{rahimi2007random}, and a single fully connected linear classification layer based on learned filters \cite{bottou2010large}.

\subsection{CNN-based feature extractor}\label{sec:tree_cnn}
We briefly review the tree-like CNN-based feature extractor presented in \cite{wiatowski2016}, the basis of which is a convolutional transform followed by a non-linearity and a pooling operation. 
Specifically, every network layer---specified by the layer index $1\leq d \leq D$---is associated with 
\begin{itemize}
\item[i)] a collection of pre-specified filters $\{ \chi_d \}\cup\{ g_{\la_d} \}_{\la_d \in \Lambda_d} \subseteq \R^{N_d\times N_d}$, indexed by a countable set $\Lambda_d$ and satisfying the Bessel condition 
\begin{equation}\label{eq:frame}
 \| f\ast \chi_d\|^2_2 \;+\! \sum_{\lambda_d \in \Lambda_d}\| f\ast g_{\lambda_d}\|_2^2\leq B_d\| f\|_2^2,\vspace{-0.2cm}
\end{equation} for all $f \in \R^{N_d\times N_d}$, for some $B_d>0$, where $\ast$ denotes the circular convolution operator,
\item[ii)] a pointwise Lipschitz-continuous non-linearity $\rho_d:\R \to \R$, \item[iii)] a Lipschitz-continuous pooling operator $P_d:\R^{N_d\times N_d} \allowbreak \to \R^{(N_d/S_d)\times (N_d/S_d)}$, where the integer $S_d \in \mathbb{N}$, with $N_d/S_d=:N_{d+1} \in \mathbb{N}$,  is referred to as pooling factor, and determines the ``size'' of the neighborhood values are combined in. \end{itemize}
 Associated with these filters, non-linearities, and pooling operators,  the feature maps are defined as  \begin{equation}\label{eq:featmap}
 f_{q_d}:=u_{q_d}\ast \chi_{d+1}\in \R^{N_{d+1}\times N_{d+1}},
\end{equation}
where $q_d=(\lambda_1,\lambda_{2},\dots, \lambda_d) \in \Lambda_1\times \Lambda_{2}\times \dots \times \Lambda_d=:\Lambda_1^d$ is a path of indices of length $d$, and 
$$
u_{q_d}=P_d\big(\rho_d(u_{q_{d-1}}\ast g_{\lambda_d})\big),
$$
for $d\in \{1,\dots D-1\}$, are propagated signals with input signal $u_{q_0}:=f\in \R^{N_1\times N_1}$. 
 \begin{remark}
 It is shown in \cite[Thm. 2]{wiatowski2016} that the feature maps \eqref{eq:featmap} are vertically translation-invariant in the sense of the layer index $d$ determining the extent to which the feature maps are translation-invariant, and deformation insensitive w.r.t. to small non-linear deformations. We refer the reader to \cite{wiatowski2015mathematical,Bruna,MallatS} for similar translation-invariant and deformation-insensitive tree-like CNNs.\end{remark}
 
To leverage the network-based feature maps \eqref{eq:featmap} for semantic segmentation, we now bi-linearly interpolate the feature maps \eqref{eq:featmap} to the size of the input image $f$, i.e., $N_1\times N_1$, according to
$
U_d f_{q_d} \in  \R^{N_{1}\times N_{1}},
$
where $U_d: \R^{N_{d+1}\times N_{d+1}}\to  \R^{N_{1}\times N_{1}}$ is the corresponding bi-linear interpolation operator with $U_0f=f$. The feature vector $F_{i,j}$ of the pixel $f_{i,j}\in \mathbb{R}$ (of the input image $f\in \R^{N_1\times N_1}$) is defined as 
\begin{equation}\label{ST}
F_{i,j}:=\Big\{(U_d f_{q_d})_{i,j}\Big\}_{q_d \in \cup_{k=0}^{D-1}\Lambda_1^k},
\end{equation}
i.e., $F_{i,j}$ is obtained by collecting the entry at location $(i,j)$ from each (up-sampled) feature map in the network (see Fig.~\ref{fig_net}). 

For the experiments in the present paper,
we particularize the feature extractor as follows: In every network layer, we employ tensorized (i.e., separable) Haar wavelets $\{ g_\la\}_{\la \in \Lambda}$, sensitive to $R = 3$ directions (horizontal, vertical, and diagonal) and---for each direction---sensitive to $J=4$ scales, with corresponding wavelet low-pass filter $\chi$, 
together satisfying 
the Bessel condition \eqref{eq:frame} with $B_d = 1$, see \cite{MallatW}.\footnote{The choice of this  wavelet type is motivated by computational efficiency (see Sec. \ref{sec:effimpl}). We did not explore other wavelet types, in particular not directional wavelets.}   
We set $D=2$ and employ the modulus non-linearity $\rho=| \cdot |$ as well as pooling by sub-sampling where we retain every second pixel, i.e., $P:\R^{N\times N}\to \R^{N/2\times N/2}$ with $(Pf)_{i,j}=f_{2i,2j}$. Note that pooling increases the robustness of the feature vector w.r.t. non-linear deformations \cite{wiatowski2016} and hence allows our architecture to deal with variation in appearance of the semantic categories.
Finally, similarly to \cite{farabet2013learning}, we also employ a variant of our network that extracts features at multiple image scales $\{s_\ell\}_{\ell=1}^L \subset \mathbb{N}$ by applying the feature extractor to multiple sub-sampled versions $\{ f^\ell \}_{\ell=1}^L$ of the input image $f$, i.e., $(f^\ell)_{i,j}=f_{s_\ell i,s_\ell j}$. We then bi-linearly interpolate the resulting feature maps to the size $N_1\times N_1$ of the input image $f$ and, for every pixel, concatenate the feature vectors from all scales into a single feature vector accordingly.

\subsection{RBF kernel approximation layer}
In the $(D+1)$-st layer of our CNN, we map the feature vectors $F_{i,j}\in \R^m$ to randomized feature vectors $\phi(F_{i,j})\in \R^{\widetilde{m}}$ with $\widetilde{m}> m$, such that $\langle \phi(F_{i,j}), \phi(F_{i',j'})\rangle \approx k(F_{i,j},F_{i',j'})$, where 
$k(F_{i,j},F_{i',j'}):=\exp(-\gamma \|F_{i,j}-F_{i',j'}\|_2^2)$ is a RBF kernel with parameter $\gamma$. In large-scale classification tasks as the one considered here, such randomized feature vectors combined with a linear classifier (see Sec. \ref{subsec:linclas}) typically allow for much faster training and inference than non-linear kernel-based classifiers \cite{rahimi2007random}. 
We follow the construction given in \cite{rahimi2007random}: Let $G\in \R^{\widetilde{m}\times m}$ be a pre-specified matrix with i.i.d. standard normal entries, and let $b\in \R^{\widetilde{m}}$ be a pre-specified vector with entries i.i.d. uniform on $[0,2\pi]$. We define the transformed feature vectors as  
\begin{equation}\label{eq:rbf_sam}
\phi(F_{i,j}):=\sqrt{\frac{2}{\widetilde{m}}}\cos\big(\gamma G F_{i,j}+b\big)\in \R^{\widetilde{m}},
\end{equation}
where $\cos(v)$, for $v\in \R^{\widetilde{m}}$, refers to element-wise application of $\cos$. We note that the RBF kernel approximation can be interpreted as a single fully connected layer with random filters and $\cos$ non-linearity (see Fig. \ref{fig_net}).

\subsection{Linear classification layer}\label{subsec:linclas}
In the last, i.e., $(D+2)$-nd layer of our CNN, we employ a linear classifier, shared across all pixels, meaning that  we apply the same classifier to all $\phi(F_{i,j})$. 
Formally, we apply a matrix $W\in \R^{K \times \widetilde{m}}$, add a bias vector $v\in \R^{K}$ according to 
\begin{equation}\label{eq:svm}
y_{i,j}:=W\phi(F_{i,j})+v\in \mathbb{R}^K,
\end{equation}
and determine the class label via a one-versus-the-rest scheme as $\arg\max_{k \in \{1, \ldots, K\}} (y_{i,j})_k$. The matrix $W$ and the vector $b$ are learned by minimizing the hinge loss with an $\ell_2$-regularization term using SGD \cite{bottou2010large}, i.e., we learn the last layer (see Fig. \ref{fig_net}) using a support vector machine (SVM).

\rev{\subsection{Relation to prior work}
Although the tree-like feature extractor from \cite{wiatowski2016} and those described in \cite{wiatowski2015mathematical,Bruna,MallatS} were employed previously for different computer vision applications, to the best of our knowledge we are the first to use this type of feature extractor for semantic segmentation. A key difference to previous applications, which use the union of all feature maps to to characterize the entire image, is the aggregation of features across feature maps to characterize individual pixels. This is also in contrast to conventional pixel classification networks \cite{farabet2013learning, pinheiro2014recurrent, paisitkriangkrai2015effective, volpi2016dense} which (unlike tree-like architectures) recombine feature channels in each layer and generate pixel-wise features only in the last layer. Our architecture is hence not a special case of pixel classification networks. Finally, a similar concept of aggregating pixel-wise features across feature maps was proposed in \cite{hariharan2015hypercolumns} for simultaneous detection and segmentation (a computer vision task that differs significantly from the semantic segmentation task considered here). However, \cite{hariharan2015hypercolumns} requires a pre-trained detection network and end-to-end training, while here we rely on pre-specified wavelet filters as well as pre-specified random filters and train the last classification layer only.}

\rev{\subsection{Efficient implementation and storage requirements} \label{sec:effimpl}
Even though we leave a highly optimized implementation of the proposed architecture for future work, we discuss possible optimizations and resulting gains. 

CNN-based semantic segmentation implementations typically spend around 90\% of the overall execution time computing the feature maps before applying pixel-wise classification. During feature extraction the vast majority of of computation effort is incurred by the convolution layers \cite{cavigelli2015accelerating}, with most CNNs using in the order of 10\,GOp/image \cite{cavigelli2015accelerating,long2015fully}, where one operation (Op) is an addition or a multiplication. 

Our feature extractor only performs convolutions with tensorized (separable) Haar wavelets, which can be implemented efficiently with the \textit{algorithme \`a trous} \cite[Sec.~5.2.2]{MallatW}. For a 1D Haar wavelet transform, $2NJ$ add/subtract operations are required for a length-$N$ signal, for $J$ scales. For the separable 2D case, the filtering is first applied in horizontal direction before applying it to the two resulting signals in vertical direction \cite[Fig.~``Decomposition Step'']{matlabSWT2}, and then repeating for the remaining scales using a strided convolution pattern, resulting in a total of $2(W+2H)J$ operations. Performing the $\chi$ filter operations before interpolating and aggregating the last layer's feature maps (which are not included in the 2D stationary wavelet transform (SWT)) requires 
$\frac{1}{2}WHJ(R^2J^2)$
more additions. For all 2D SWT convolution operations in the feature extractor the number of additions amounts to $6WHJ + \frac{3}{2}WHRJ^2$ (recall that we employ pooling by subsampling). Additionally,
$WHRJ(1+RJ/4)$ 
absolute-value operations are required and $WH(RJ+1)RJ$ pixels need to be interpolated with 4 multiply and 4 add operations each. For the configuration specified in Sec.~\ref{sec:tree_cnn} applied to each of the three channels of the input image, the evaluation of the feature extraction network is dominated by the interpolation effort and results in 
387\,MOp for a $320\times 240$ image---about $20\times$ less computation effort than competing CNNs and, just as important for embedded devices, without any trained parameters to be stored. 

Furthermore, note that the evaluation of the RBF approximation layer can be accelerated from $\mathcal O(\tilde m m)$ operations to $\mathcal O(\tilde m \log m)$ operations by using a fast RBF kernel approximation such as \cite{le2013fastfood, choromanski2016recycling}. These methods (just like the one used for our evaluations) only require randomly generated values, which can be regenerated on-the-fly using an identically seeded pseudo-random number generator. 

The only parameters to be stored are the matrix $W$ in the final linear layer and the corresponding bias vector $v$. Our architecture thus has an extremely small memory footprint of only $(\tilde m+1)K$ values, which typically amounts to a few tens of thousands of parameters---a strong requirement for performing inference on tightly resource-bound hardware platforms. 
In contrast, many popular (pre-trained) CNN architectures such as AlexNet, VGG, or GoogLeNet, which the semantic segmentation CNNs in \cite{long2015fully, eigen2015predicting,  liang2015semantic, marmanisa2016semantic, sherrah2016fully} build upon, have millions of parameters, even when truncated \cite{han2015learning}.

}

\section{Experiments} \label{sec:exp}

We evaluate the performance of the proposed CNN architecture in semantic segmentation of outdoor scene images and aerial images. For both segmentation tasks 
we set the output dimension of the RBF approximation layer to $\tilde m = 5000$\rev{, resulting in a model size of less than 350kByte in all cases (for model parameters with 32bit float precision, stored using Python built-in serialization via \texttt{pickle})}. 
For training, we first randomly draw a subset containing 2\% of all pixels in the training set, collect the corresponding feature vectors, and then perform SGD passes over randomly shuffled versions of the so-obtained feature vector set.
Furthermore, we tune the RBF kernel parameter $\gamma$ as well as the parameter $\lambda$ balancing the hinge loss and the regularization term in the SVM objective (see \cite[Tab. 1]{bottou2010large}). We evaluate the segmentation performance for feature extraction at a single scale (i.e., the original image size) and at scales $s_\ell \in \{1,2,4\}$. Preliminary experiments showed that using the full training set for the SGD passes or increasing the feature extraction network depth $D$ does not significantly increase the accuracy. 
We implemented the proposed architecture in Python and Matlab (for the feature extractor) on CPU. Note that we did not optimize the implementation for speed. The runtimes we report were obtained on a desktop computer with 2.5 GHz Intel Core i7 (I7-4870HQ) and 16 GB RAM, and can be drastically reduced by leveraging the parallel processing capabilities of GPUs for the the evaluation of the feature extractor, the random layer, and the classification layer. 
To ensure a fair comparison with accuracies reported in the literature for other architectures, we always refer to the accuracies obtained without post-processing (using, e.g., a CRF or superpixels).

\subsection{Outdoor scene semantic segmentation}
\label{sec:scenelab}

We use the Stanford Background data set \cite{gould2009decomposing} containing 715 RGB images of outdoor scenes of size approximately $320 \times 240$ pixels. Each pixel is labeled with one of eight semantic categories (``sky'', ``tree'', ``road'', ``grass'', ``water'', ``building'', ``mountain'',  and ``foreground object''). 
An image is processed by first transforming it to YUV color space, applying the feature extraction network (possibly at multiple scales) to each color channel, and subsequently concatenating the extracted feature vectors. This results in feature vectors $F_{i,j}$ of dimension $m = 309$ for 1
scale and $m = 927$ for 3 scales.
The runtime per image for segmentation was $9.6$s and $23.4$s for 1 scale and 3 scales, respectively.

Fig. \ref{fig:nimg} (b, c) shows example outputs of our CNN for images from the Stanford Background data set. Following \cite{farabet2013learning}, we estimate the accuracy of our method using 5-fold cross-validation (CV). Table \ref{tab:outaerialimg} (top) shows the pixel accuracy (averaged over all pixels) and the class accuracy (i.e., the average class precision) of our CNN architecture, along with the end-to-end trained pixel classification CNN from \cite{farabet2013learning} (we refer to \cite[Tab. 1]{farabet2013learning} for comparison with non-CNN-based methods). For 1 scale, our network outperforms the CNN from \cite{farabet2013learning}. Using 3 scales instead of 1 increases the pixel accuracy of our CNN by 3.4\%, i.e., the gain from using multiple scales is smaller than for the CNN from \cite{farabet2013learning}. A possible reason for this could be that the wavelet filters used in our network already capture the multi-scale nature of the images quite well. \rev{We note that other CNN architectures \cite{pinheiro2014recurrent, cavigelli2015accelerating, long2015fully, eigen2015predicting, liang2015semantic, badrinarayanan2015segnet} achieve higher accuracies in semantic segmentation of outdoor scenes (mostly on other data sets). However, these architectures are all trained end-to-end and rely either on deconvolution layers or recurrent structure, resulting in orders of magnitude larger models and requiring considerably larger computational effort for training. In addition, the architectures from \cite{long2015fully, eigen2015predicting, liang2015semantic, badrinarayanan2015segnet} are based on pre-trained classification CNNs.}

\paragraph*{Effect of training set size on pixel accuracy} We investigate the effect of the number of training images $n_\mathrm{train}$ on the pixel accuracy. Specifically, we fix $\gamma$ and $\lambda$ to the values obtained for 5-fold CV, randomly split the data set into 500 training and 215 testing images, and draw $n_\mathrm{train}$ images from the training set (keeping the same training/testing split for all $n_\mathrm{train}$).
Fig. \ref{fig:nimg} (a) shows the pixel accuracy of our architecture 
as a function of $n_\mathrm{train}$ (averaged over 3 random training/testing splits).
For $n_\mathrm{train} \geq 100$ the pixel accuracy exceeds 93\% of the accuracy obtained for the full training set containing $575$ images (for 5-fold CV), for both 1 scale and 3 scales. Our architecture is thus quite data efficient. In particular, in the single scale case, it matches the pixel accuracy of the CNN from \cite{farabet2013learning} using 5 times less training images.

\subsection{Aerial image semantic segmentation}
We evaluate our architecture on the Vaihingen data set\footnote{The Vaihingen data set was provided by the German Society for Photogrammetry, Remote Sensing and Geoinformation (DGPF) \cite{cramer2010dgpf}: \url{http://www.ifp.uni-stuttgart.de/dgpf/DKEP-Allg.html}.} of the ISPRS 2D semantic labeling contest \cite{isprsdataset}, 
which contains 33 aerial images of varying size (average size $2494 \times 2064$ pixels \cite{volpi2016dense}). Pixel-level semantic labels (categories: ``impervious surface'', ``building'', ``low vegetation'', ``tree'', ``car'', and ``background''
) are available for $16$ images, the labels of the remaining images serve as private testing set for the contest. The images have three channels (near infrared, red, and green) and come with a (coregistered) digital surface model (DSM). Following \cite{volpi2016dense, sherrah2016fully}, we retain images $11$, $15$, $28$, $30$, and $34$ for testing and use the remaining labeled images in the data set for training. Similarly as for outdoor scene semantic labeling, we apply the feature extraction network to each channel and to the normalized version of the DSM provided by \cite{gerke2015use}. We concatenate the feature vectors extracted for each channel to obtain feature vectors $F_{i,j}$ of dimension $m = 412$ for $1$ scale and $m = 1236$ for $3$ scales. The runtime per megapixel for segmentation was $2.6$\,min and $6.6$\,min for $1$ and $3$ scales, respectively.

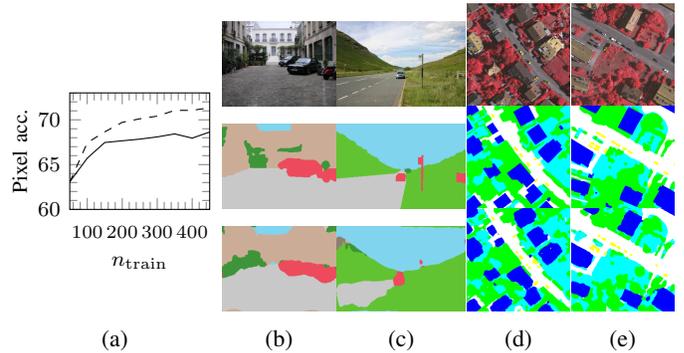
\begin{figure}\hspace{-3mm}
    \begin{subfigure}[b]{0.35\columnwidth}
        \centering
        \begin{tikzpicture}[scale=1] 
            \begin{axis}[
            scale only axis,
            ylabel=Pixel acc.,
            xlabel=$n_\mathrm{train}$,
            width = 0.6\columnwidth,
            height = 0.5\columnwidth,
            ymin = 60,
            ymax = 73,
            xmin = 50,
            xmax = 450,
            minor tick num=4,
            xticklabel style={yshift=-1mm,font=\scriptsize},
            yticklabel style={font=\footnotesize},
            ylabel style={yshift=-1mm,font=\footnotesize},
            xlabel style={font=\footnotesize},]                 \addplot +[mark=none,solid,black] table[x index=0,y index=1]{./acc_avg_nimg_ss.dat};
                \addplot +[mark=none,dashed,black] table[x index=0,y index=1]{./acc_avg_nimg_ms.dat};
            \end{axis}
        \end{tikzpicture}
        \vspace{0.45cm}
        \caption{}
  \end{subfigure}\hspace{-0.2cm}
  \begin{subfigure}[b]{0.77\columnwidth}
        \begin{subfigure}[b]{0.22\columnwidth}
        \includegraphics[width=\columnwidth]{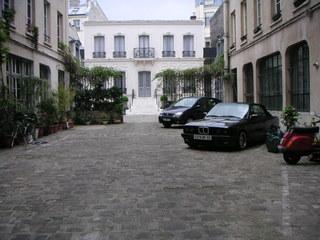}
    \end{subfigure}\hspace{-0.1cm}
    \begin{subfigure}[b]{0.25\columnwidth}
        \includegraphics[width=\columnwidth]{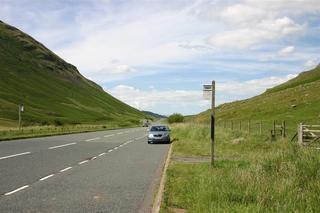}
    \end{subfigure}\hspace{-0.1cm}
    \begin{subfigure}[b]{0.2\columnwidth}
        \includegraphics[width=\columnwidth]{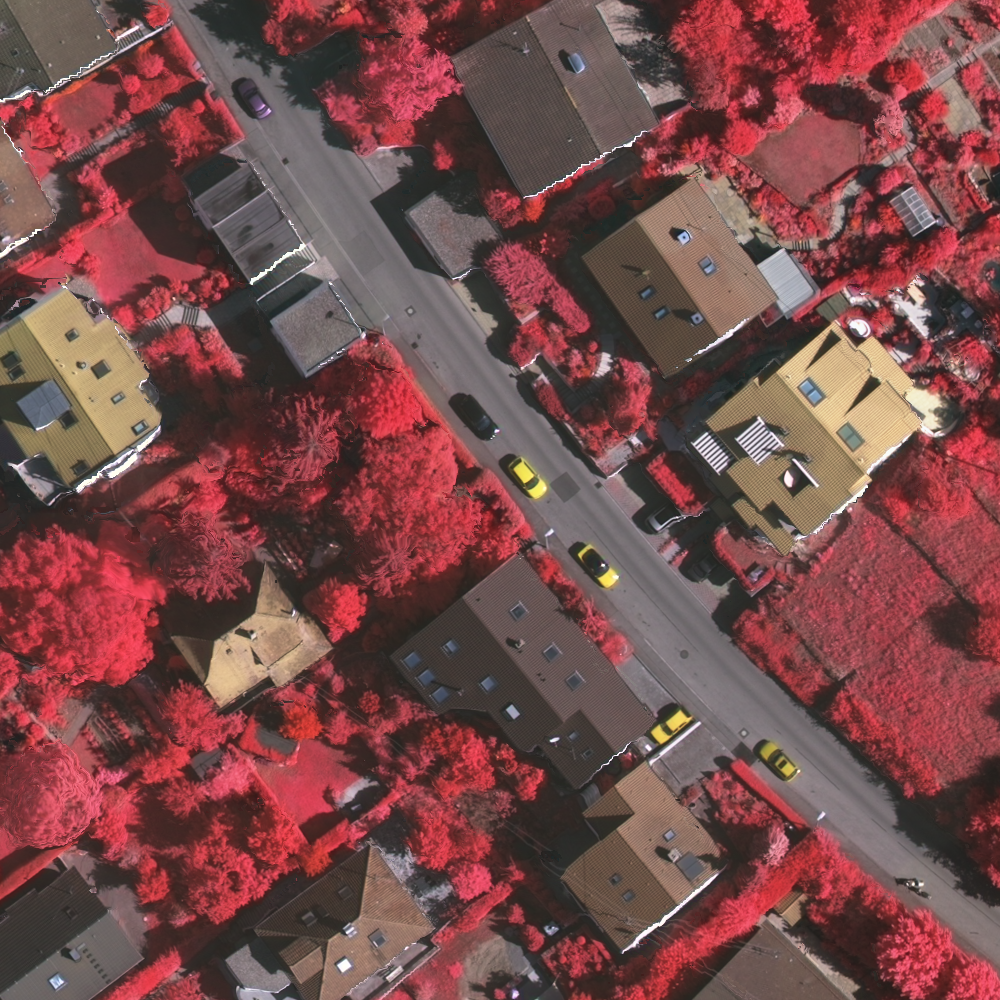}
    \end{subfigure}\hspace{-0.1cm}
    \begin{subfigure}[b]{0.2\columnwidth}
        \includegraphics[width=\columnwidth]{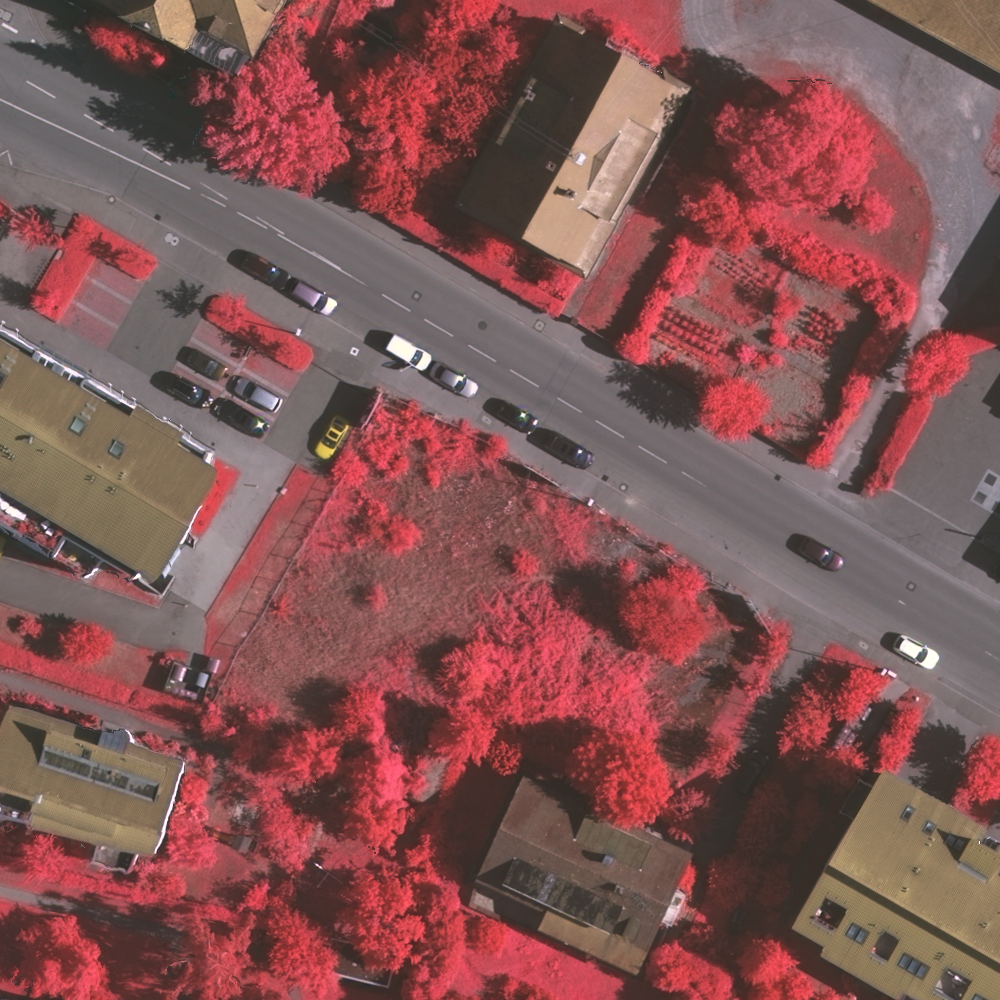}
    \end{subfigure} \\
        \begin{subfigure}[b]{0.22\columnwidth}
         \includegraphics[width=\columnwidth]{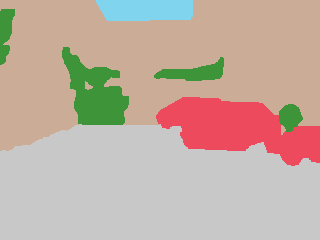}
    \end{subfigure}\hspace{-0.1cm}
    \begin{subfigure}[b]{0.25\columnwidth}
         \includegraphics[width=\columnwidth]{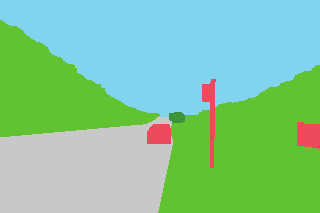}
    \end{subfigure}\hspace{-0.1cm}
    \begin{subfigure}[b]{0.2\columnwidth}
        \includegraphics[width=\columnwidth]{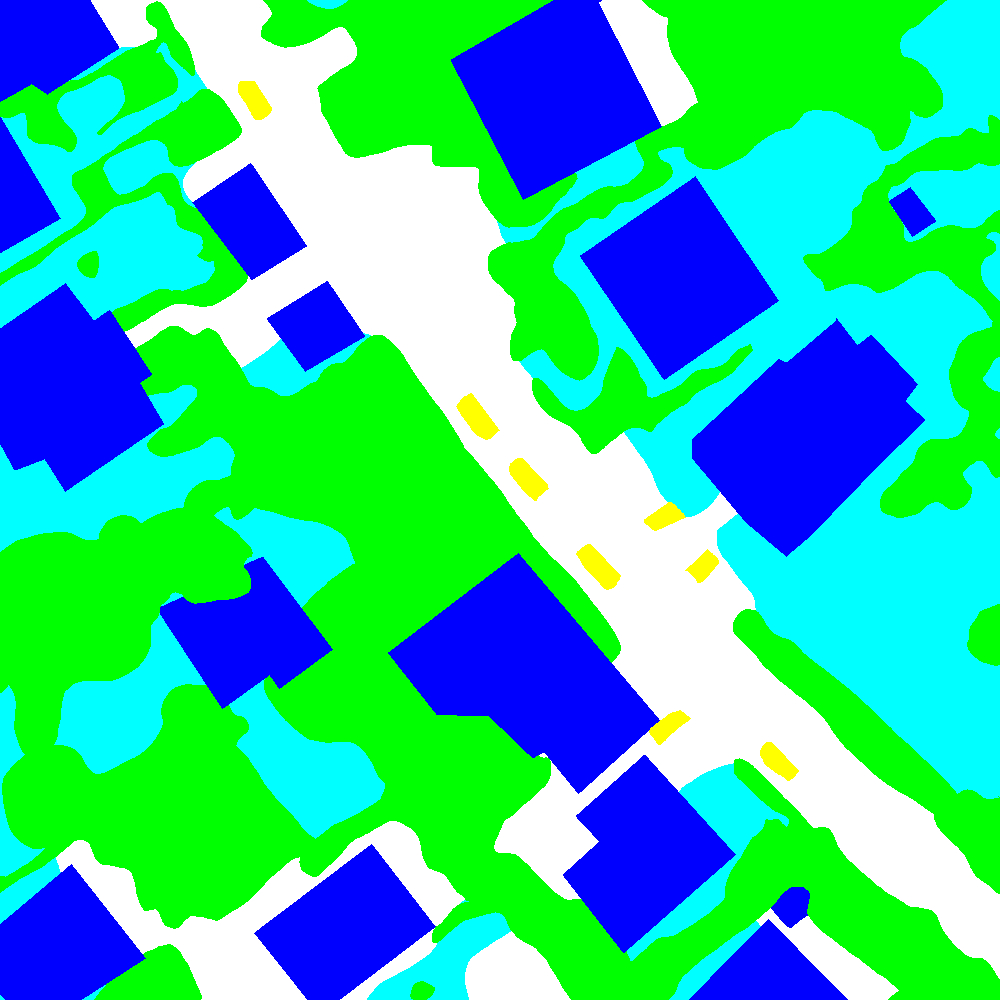}
    \end{subfigure}\hspace{-0.1cm}
    \begin{subfigure}[b]{0.2\columnwidth}
        \includegraphics[width=\columnwidth]{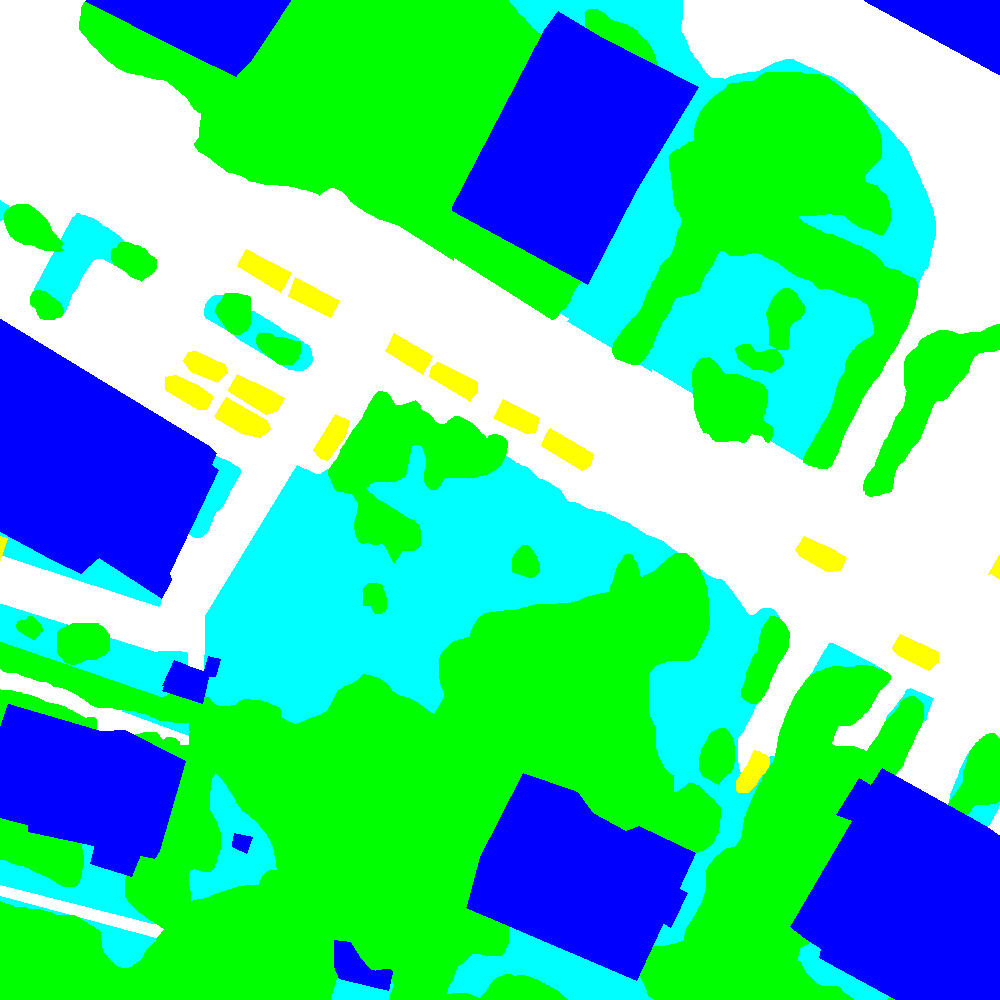}
    \end{subfigure}\\
    \begin{subfigure}[b]{0.22\columnwidth}
        \includegraphics[width=\columnwidth]{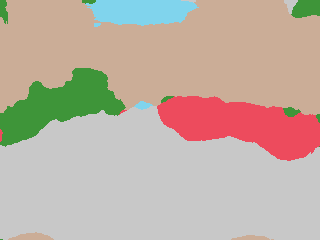}
        \caption{}
    \end{subfigure}\hspace{-0.1cm}
    \begin{subfigure}[b]{0.25\columnwidth}
        \includegraphics[width=\columnwidth]{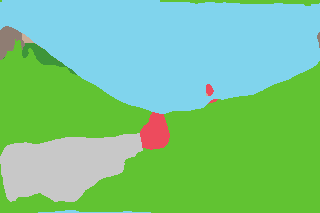}
        \caption{}
    \end{subfigure}\hspace{-0.1cm}
    \begin{subfigure}[b]{0.2\columnwidth}
        \includegraphics[width=\columnwidth]{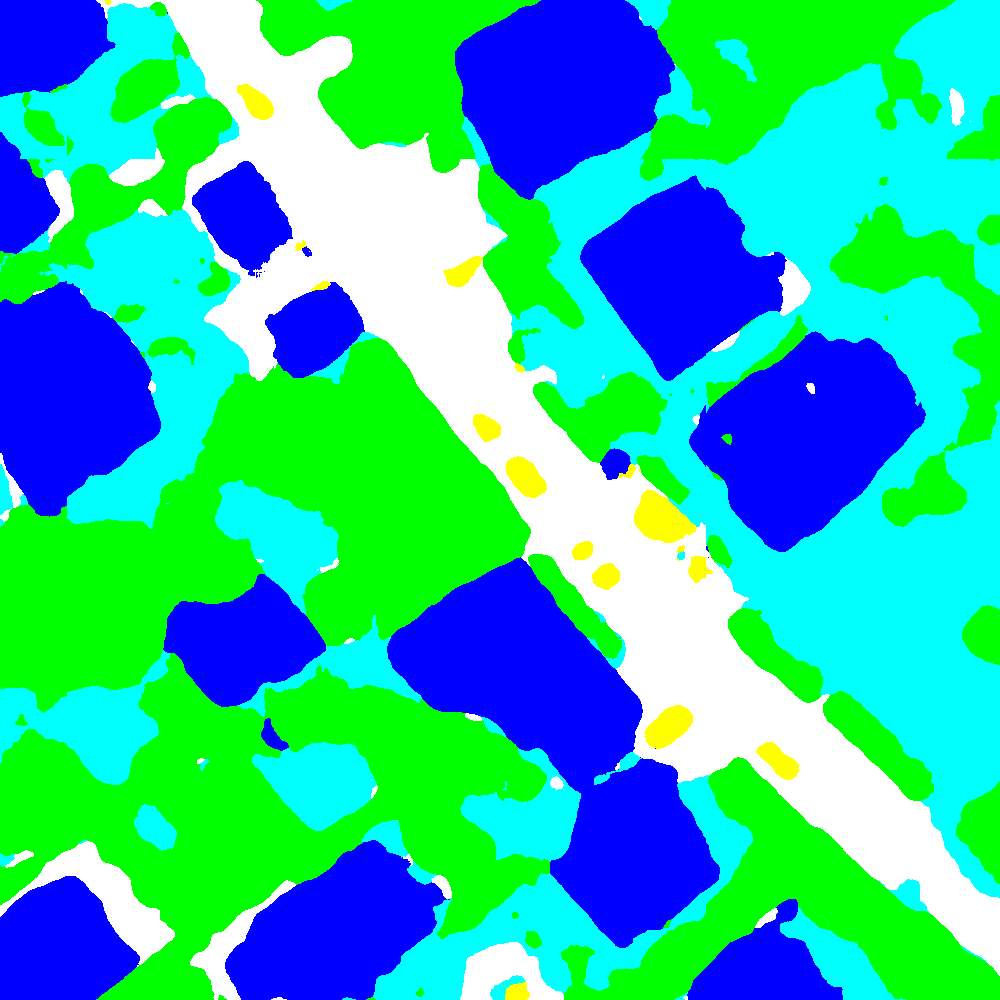}
        \caption{}
    \end{subfigure}\hspace{-0.1cm}
    \begin{subfigure}[b]{0.2\columnwidth}
        \includegraphics[width=\columnwidth]{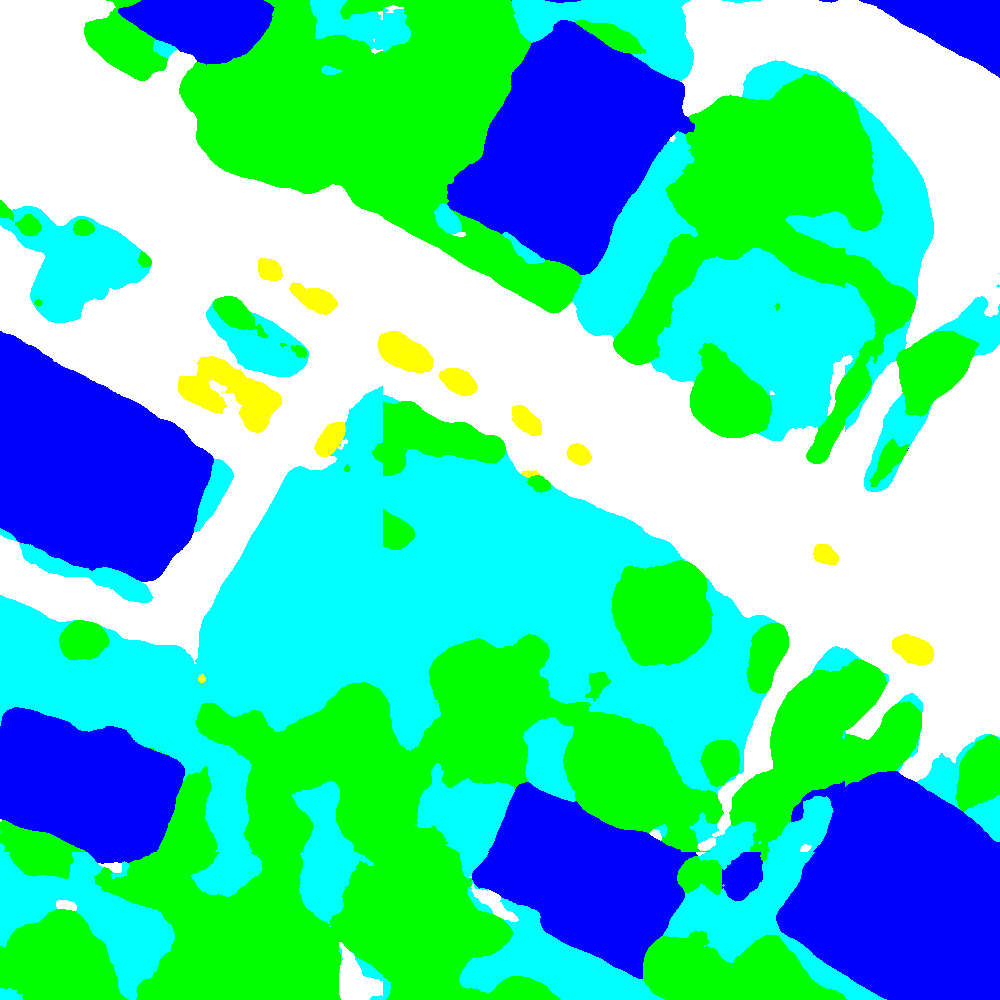}
        \caption{}
    \end{subfigure}\hspace{-0.1cm}
  \end{subfigure}\vspace{-0.1cm}
  \caption{\label{fig:nimg}(a): Pixel accuracy (in \%) of our architecture as a function of the number of training images $n_\mathrm{train}$ using 1 scale (solid) and 3 scales (dashed), for outdoor scene semantic segmentation. (b--e): Input image (top), ground truth labeling (middle), and prediction by our architecture (3 scales; bottom) for example outdoor scenes (b, c) and aerial images (d, e).}
\vspace{-0.4cm}
\end{figure}

Fig. \ref{fig:nimg} (d, e) show an example of an aerial image patch along with the ground truth labeling and the predictions of our architecture. In Table \ref{tab:outaerialimg} (bottom) we report the pixel accuracy and the F1 score (averaged over classes) of our CNN architecture as well as that of two end-to-end trained pixel classification CNNs \cite{paisitkriangkrai2015effective,volpi2016dense}. As in \cite{paisitkriangkrai2015effective,volpi2016dense} (and following the rules of the ISPRS 2D semantic labeling contest) labeling errors within a 3 pixel radius of the (true) category boundaries are excluded for the computation of the accuracy and the F1 score. It can be seen that the accuracy and the average F1 score of our architecture is \rev{comparable to} the pixel classification CNNs proposed in \cite{paisitkriangkrai2015effective,volpi2016dense}.
On the private ISPRS testing set our architecture with 3 scales obtained a pixel accuracy of 85.9 \%, thus outperforming the algorithms from \cite{lagrange2015convolutional}, each of which combines a pre-trained classification CNN with a SVM. 
We note that other CNN architectures \cite{paisitkriangkrai2015effective,marmanisa2016semantic,volpi2016dense,sherrah2016fully} achieve higher accuracies and F1 scores on the Vaihingen data set. Again, these architectures are all trained end-to-end and rely on deconvolution layers \cite{volpi2016dense, marmanisa2016semantic}, pre-trained classification CNNs \cite{marmanisa2016semantic,sherrah2016fully}, CRF-based refinement \cite{paisitkriangkrai2015effective, sherrah2016fully}, and/or additional hand-crafted features \cite{paisitkriangkrai2015effective}.

\begin{table}
\rev{
\centering
\normalsize{
\begin{tabular}{ c l c c c c}
 \multicolumn{1}{c}{} & & \multicolumn{2}{c}{Pixel acc.} & \multicolumn{2}{c}{Class acc.} \\
 \multicolumn{1}{c}{} & Number of scales & 1 & 3 & 1 & 3 \\
  \hline
  \multirow{2}{*}{\rotatebox[origin=c]{90}{out.}} & Pixel class. CNN \cite{farabet2013learning} & 66.0 & 78.8 & 56.5 & 72.4 \\
  \cline{2-6}
  & Our CNN & 68.3 & 71.7 & 62.1 & 63.3  \\
  \hline \\[-0.3cm]
  \multicolumn{1}{c}{} & & \multicolumn{2}{c}{Pixel acc.} & \multicolumn{2}{c}{F1 score} \\
    \hline
  \multirow{3}{*}{\rotatebox[origin=c]{90}{aerial}} & Pixel class. CNN \cite{paisitkriangkrai2015effective} & 83.46 & 85.56 & 77.84 & 81.74 \\
    & CNN-PC \cite{volpi2016dense} & 86.67 & - & 68.01 & - \\   \cline{2-6}
  & Our CNN & 85.16  & 85.45 & 66.28 & 75.73 \\
  \hline
\end{tabular}}
}
\caption{\label{tab:outaerialimg} \rev{Pixel accuracy (in \%) and average Class accuracy/F1 score (in \%) for outdoor scene (top, ``out.'') and aerial image (bottom, ``aerial'') semantic segmentation on the test set.}}\vspace{-0.4cm}
\end{table}

\section{Conclusion}
\rev{We proposed a simple highly structured Haar wavelet-based CNN architecture for semantic segmentation with an extremely small model size, potentially allowing for a very fast implementation thanks to its structure. We demonstrated that our architecture is very data efficient and that its accuracy is comparable to that of conventional pixel classification CNNs, even though we used pre-defined features in place of computationally demanding end-to-end feature learning.} 

Replacing the pixel classification network by deconvolution layers might improve the segmentation accuracy and is an interesting direction to be explored in the future.

\section*{Acknowledgments}
The authors would like to thank J.~K{\"u}hne for preliminary work on the experiments in Section~\ref{sec:scenelab} and M.~Lerjen for help with computational issues. L. Cavigelli and L. Benini gratefully acknowlege funding by armasuisse Science \& Technology.

\bibliographystyle{IEEEbib_abbr}

\bibliography{semanticlabeling}

\end{document}